\documentclass[10pt,twocolumn,letterpaper]{article}

\usepackage[pagenumbers]{cvpr} 

\usepackage{graphicx}
\usepackage{float}
\usepackage{multicol}
\usepackage{amssymb}
\usepackage{algorithm}
\usepackage{algorithmic}
\usepackage{xcolor}
\usepackage{array}
\usepackage{tabularx,colortbl}
\usepackage{rotating}
\newcommand{\first}[1]{\textbf{\textcolor{red}{#1}}}

\newcommand{\second}[1]{\textbf{\textcolor{blue}{#1}}}

\newcommand{\INPUT}{\item[\textbf{Input:}]}
\newcommand{\OUTPUT}{\item[\textbf{Output:}]}
\newcommand{\METHODNAME}{Grounded-Dreamer }

\newcommand{\mypara}{\vspace*{-4mm}\hspace{0mm}\paragraph}

\definecolor{cvprblue}{rgb}{0.21,0.49,0.74}
\usepackage[pagebackref,breaklinks,colorlinks,citecolor=cvprblue]{hyperref}

\def\paperID{12689} 
\def\confName{CVPR}
\def\confYear{2024}

\title{Grounded Compositional and Diverse Text-to-3D with Pretrained Multi-View Diffusion Model}
\author{Xiaolong Li  
\qquad Jiawei Mo  
\qquad Ying Wang   
\qquad Chethan Parameshwara
\qquad Xiaohan Fei   \\
\qquad Ashwin Swaminathan   
\qquad CJ Taylor  
\qquad Zhuowen Tu  \\
\qquad Paolo Favaro  
\qquad Stefano Soatto \\
AWS AI Labs\\
{\tt\small lxiaolx@amazon.com}}

\begin{document}
\makeatletter
\g@addto@macro\@maketitle{
\vspace{-6mm}
\begin{figure}[H]
\setlength{\linewidth}{\textwidth}
\setlength{\hsize}{\textwidth}
    \includegraphics[width=1.0\linewidth]{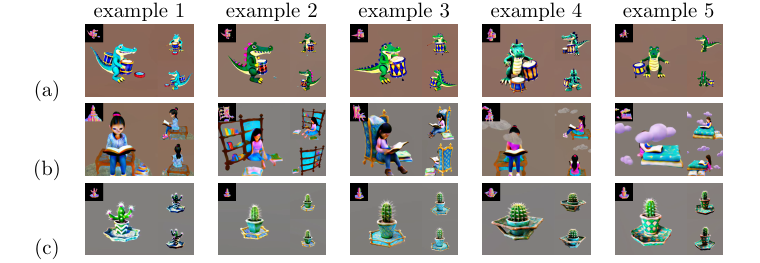}
    \caption{\textit{Diverse 3D assets generated with \METHODNAME}. Text prompts: (a) “a crocodile playing a drum set”, (b) “A girl is reading a hardcover book in her room”, (c) “a green cactus in a hexagonal cup on a star-shaped tray”.}
    \label{fig:t3d_diverse}
\end{figure}
}
\maketitle

\begin{abstract}
In this paper, we propose an effective two-stage approach named \METHODNAME to generate 3D assets that can accurately follow complex, compositional text prompts while achieving high fidelity by using a pre-trained multi-view diffusion model. Multi-view diffusion models, such as MVDream\cite{shi2023mvdream}, have been shown to generate high-fidelity 3D assets using score distillation sampling (SDS). However, applied naively, these methods often fail to comprehend compositional text prompts, and may often entirely omit certain subjects or parts. To address this issue, we first advocate leveraging text-guided 4-view images as the bottleneck in the text-to-3D pipeline. We then introduce an attention refocusing mechanism to encourage text-aligned 4-view image generation, without the necessity to re-train the multi-view diffusion model or craft a high-quality compositional 3D dataset. We further propose a hybrid optimization strategy to encourage synergy between the SDS loss and the sparse RGB reference images. Our method consistently outperforms previous state-of-the-art (SOTA) methods in generating compositional 3D assets, excelling in both quality and accuracy, and enabling diverse 3D from the same text prompt. 
\end{abstract}    
\section{Introduction}
\label{sec:intro}



The quest to transform textual descriptions into vivid 3D models has seen remarkable advancements with methods like Score Jacobian Chaining \cite{wang2023score}, DreamFusion \cite{poole2023dreamfusion}, and subsequent developments like \cite{lin2023magic3d, chen2023fantasia3d, shi2023mvdream}. However, the field still faces significant challenges in accurately rendering compositional prompts and ensuring diversity in the synthesized objects. Our research introduces a novel approach that not only addresses these challenges but also represents a paradigm shift in text-to-3D synthesis.

We draw inspiration from the 2D domain, where the same pre-trained diffusion models can generate compositionally correct images under multiple attempts. These images can serve as robust references for 3D synthesis, leading us to the key question: can we leverage these text-conditioned, diverse, compositionally correct views to enhance 3D asset creation? A naive solution is to combine text-to-image (T2I) and single-image-to-3D pipelines, such as \cite{qian2023magic123,liu2023syncdreamer, shi2023zero123++, long2023wonder3d}. However, they often result in inconsistent geometry or semantics mainly due to inherent ambiguities and the domain gap when conditioning on a single image. This inconsistency is particularly evident in 3D assets where different views (front, side, rear) appear incongruent. 

In our approach, we advocate for establishing a more robust foundation for Text-to-3D synthesis by utilizing multi-view images. Instead of relying on a single view, which often leads to ambiguities and inconsistencies, we generate four spatially distinct views, each separated by 90 degrees. This multi-view approach effectively constrains and defines an object's shape and appearance, bridging the gap between 2D imagery and 3D modeling. By employing a pre-trained multi-view diffusion model \cite{shi2023mvdream}, we can generate these four views from a text prompt in a multi-view consistent manner. This process of generating and utilizing multiple views provides a more reliable and ``grounded" basis for 3D reconstruction, as it reduces the uncertainty often associated with interpreting and extrapolating from a single image.

However, akin to the limitations of Stable Diffusion \cite{rombach2022high-ldm} in generating compositional single-view images \cite{huang2023t2i}, advanced models like MVDream can also struggle to consistently produce four-view images that accurately capture the correct compositional subjects, attributes, and their spatial relationships. To counter this,  our first stage employs an attention refocusing mechanism during the inference phase, as inspired by \cite{chefer2023attend}. This strategy ensures that each subject token from the text is precisely represented across all views, effectively addressing the ambiguities common in single-view reconstructions. By enhancing compositional accuracy without the need for re-training or fine-tuning the existing multi-view diffusion model, our method not only conserves resources but also leverages the rich knowledge embedded in the pre-trained text-guided diffusion model. This approach promotes greater adaptability and creativity in a wide range of scenarios.

In the second stage of our method, we implement a nuanced, coarse-to-fine reconstruction process. This stage is characterized by an integration of sparse-view Neural Radiance Fields (NeRF) with text-guided diffusion priors. The process begins by establishing a coarse 3D structure using sparse-view NeRF, grounded in the compositional accuracy achieved in the first stage. We then refine the details of this structure by introducing text-guided diffusion priors. A critical component of this stage is the implementation of a delayed Score Distillation Sampling (SDS) loss, coupled with an aggressively annealed timestep schedule. This combination is designed to refine textures and geometries in a scene-agnostic manner, ensuring that the enhancements do not distort the compositional accuracy established earlier.

\begin{figure*}[ht]
    \centering
    \begin{subfigure}{.45\textwidth}
        \centering
        \includegraphics[width=\linewidth]{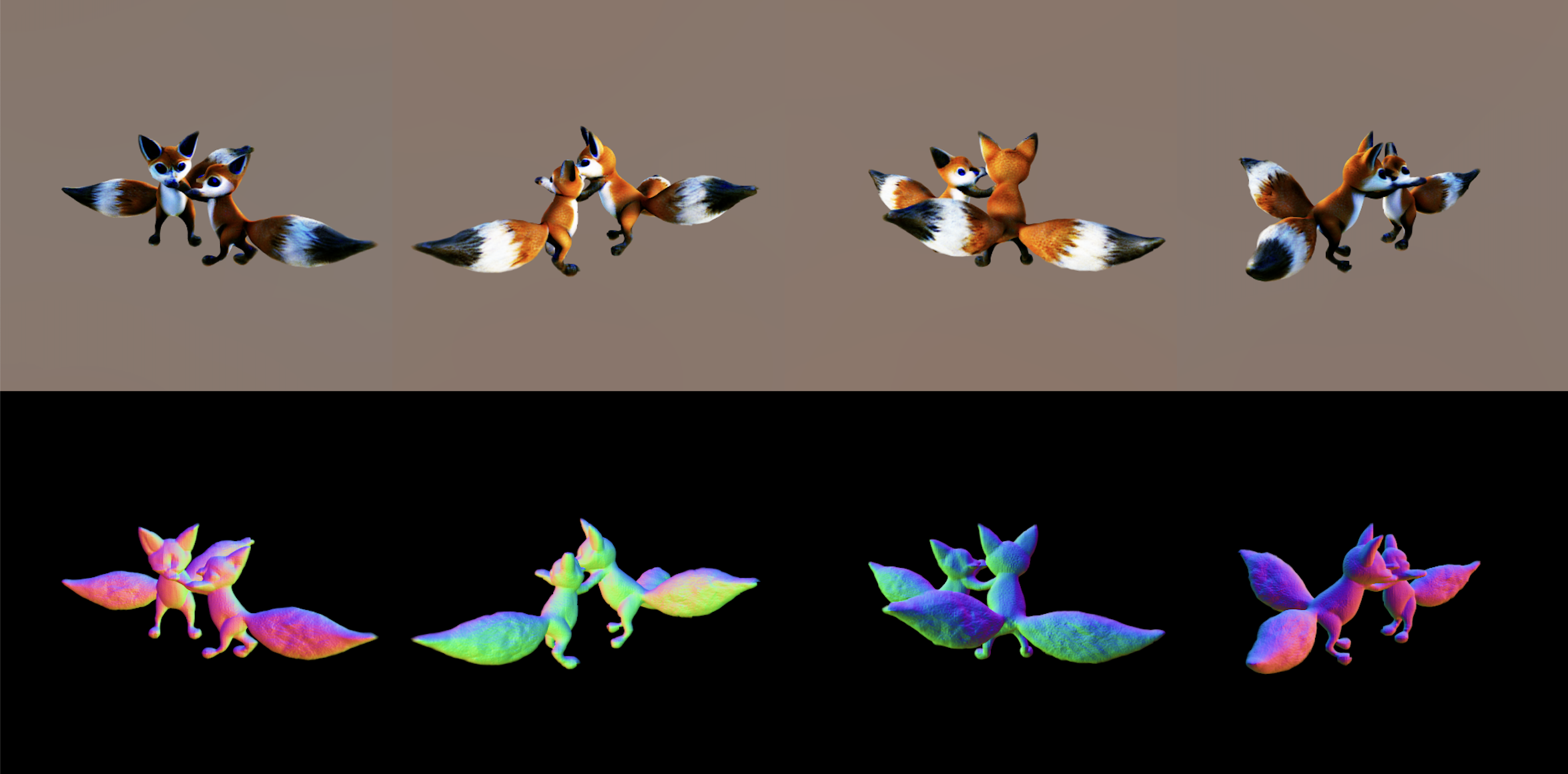}
        \caption{Running into the Janus issue}
    \end{subfigure}%
    \hfill
    \begin{subfigure}{.45\textwidth}
        \centering
        \includegraphics[width=\linewidth]{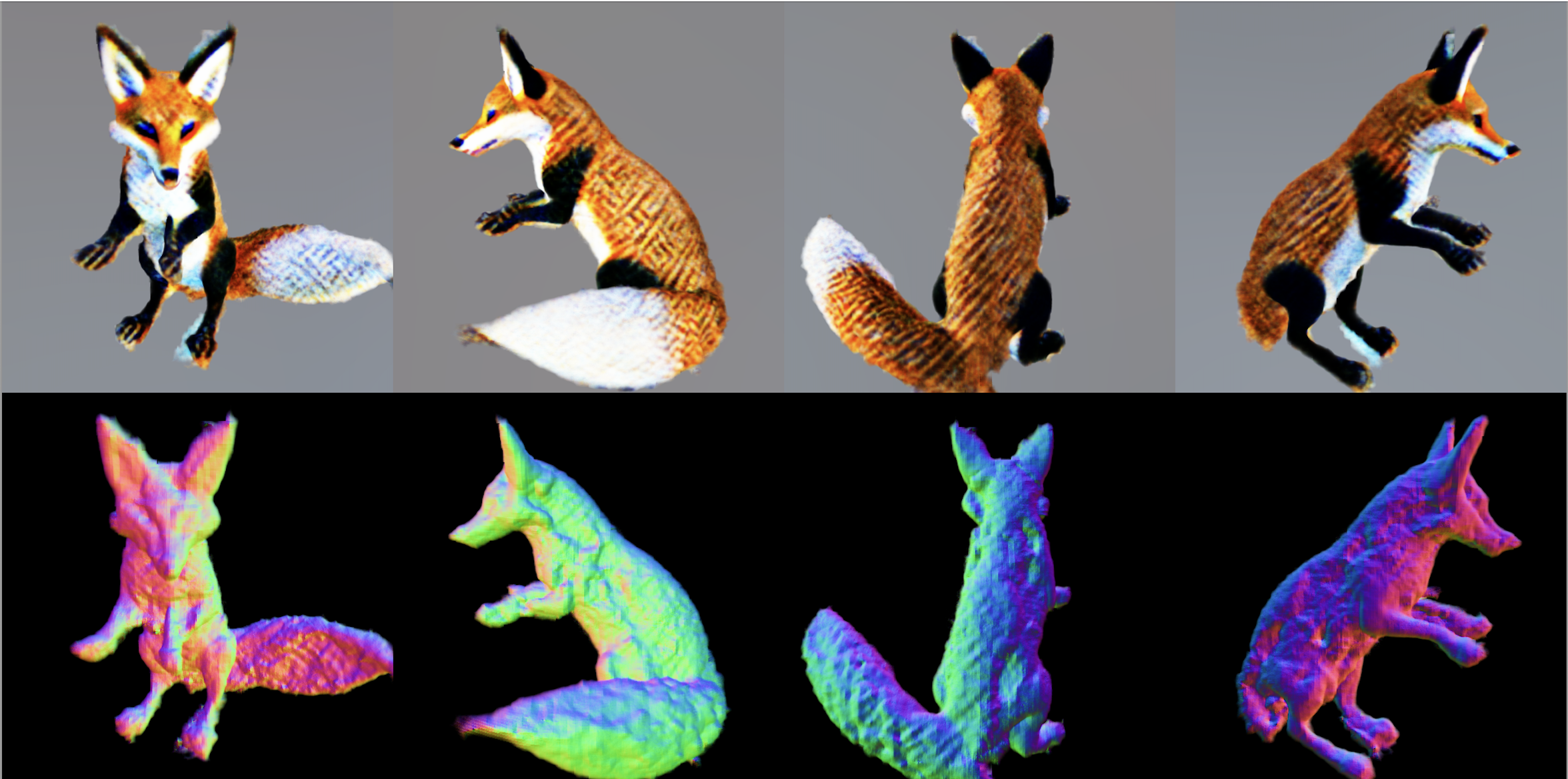}
        \caption{Ignoring compositional priors}
    \end{subfigure}%
    \captionof{figure}{\textit{Illustration of common failure patterns when naively combining sparse reference images to the SDS loss}. Text prompt: ``Two foxes fighting''. When combining 4 reference images, (a) ends up generating a fox with two tails, while (b) misses the `two' information.} 
    \label{fig:common_failure}
\end{figure*}

It is important to note that a straightforward combination of sparse-view image supervision with existing Text-to-3D pipelines can lead to significant geometric distortions, such as the duplication of body parts (commonly referred to as the `Janus' issue) or a complete disregard for compositional priors, resulting in a regression to the original MVDream Text-to-3D outputs. These common failure patterns, as illustrated in \cref{fig:common_failure}, underscore the need for a more sophisticated approach to integrating these elements. Our method's staged process, with its careful balance of NeRF and diffusion priors, is designed to avoid these pitfalls, ensuring a coherent and accurate 3D representation.

By integrating our novel two-stage framework with a pre-trained multi-view diffusion model, we develop an effective pipeline for compositional Text-to-3D synthesis that accurately adheres to complex text prompts. Our method not only generates diverse 3D assets for the same text prompts by varying the sets of four-view images but also marks significant advancements in the field:
\begin{itemize}
\item
Innovative Two-Stage Framework: We introduce a new paradigm in Text-to-3D synthesis, where sparse-view images generated from a Text-to-Image (T2I) model serve as an intermediary, ensuring the preservation of compositional priors and facilitating diverse 3D generation.

\item
Compositional Alignment via Test-Time Optimization: Our method includes a novel test-time optimization technique for multi-view generation, significantly improving text-image alignment, particularly in terms of compositional accuracy.

\item
Hybrid Training Strategy for High-Fidelity 3D Assets: We propose a synergistic training approach that combines few-shot NeRF with Score Distillation Sampling (SDS)-based optimization. This strategy not only achieves high-fidelity, text-guided 3D asset generation but also maintains precise compositional relationships.
\end{itemize}

\section{Related Works}
\label{sec:related}

\subsection{Text-to-3D}
Early works like CLIP-Mesh \cite{mohammad2022clip} and DreamField first show the possibility of generating 3D assets with text prompts using 2D priors. Later Score Distillation Sampling (SDS) is proposed by Boole~\etal in DreamFusion~\cite{poole2023dreamfusion} and followed by many~\cite{wang2023sjc,lin2023magic3d,tsalicoglou2023textmesh,metzer2023latent,chen2023fantasia3d,wang2023prolificdreamer, shi2023mvdream,tang2023dreamgaussian,chen2023text-gsgen,yi2023gaussiandreamer,li2023sweetdreamer,sun2023dreamcraft3d}. The key idea is to supervise NeRF~\cite{mildenhall2021nerf, barron2021mip} training using the supervision signals from a pre-trained and frozen large text-to-image model~\cite{rombach2022high-ldm, saharia2022photorealistic_imagen}, which can be considered as distilling deterministic generators (i.e. neural radius field)
as student models for a pre-trained large-scale diffusion models conditioned on specific text \cite{luo2023comprehensive}. Several techniques have been proposed to improve the SDS framework including better viewpoint conditioning~\cite{armandpour2023perpneg}, better timestep scheduling~\cite{huang2023dreamtime}, variational score distillation~\cite{wang2023prolificdreamer}, accelerated NeRF representation~\cite{muller2022instant}, surface representation~\cite{yariv2021volume,wang2021neus, shen2021deep-dmtet}, improved efficiency using Gaussian Splatting~\cite{kerbl20233d-gaussiansplatting}, and improved fidelity~\cite{zhu2023hifa}. 
However, these methods only demonstrate limited capability of generating compositional or diverse 3D assets.
\subsection{Sparse Image-to-3D with Diffusion Models}

A number of image-to-3D methods in existing literature attempt to synthesize 3D models from a single image, including RealFusion \cite{melas2023realfusion}, Zero-1-to-3 \cite{liu2023zero1to3}, Magic123 \cite{qian2023magic123}, and Wonder3D \cite{long2023wonder3d}. Such methods are often used as part of a two-stage text-to-3D pipeline in which an image is first generated using a high quality text-to-image model, and subsequently used as a reference for 3D object synthesis. Although such approaches bear the advantage of the user being able to ``select" the desired aesthetics before 3D synthesis commences, a single image is unable to fully capture multi-object compositions in 3D space in which objects may occlude each other, or when the prompt describes details that are impossible to capture from a single camera view, and easily becomes a bottleneck in preserving 3D compositional accuracy in the full text-to-3D pipeline.

\subsection{3D Compositional Generation}
Previous works on compositional generation can be divided into two major tracks, layout or depth-conditioned 3D scene generation, like \cite{wang2021sceneformer, bautista2022gaudi, bahmani2023cc3d,wei2023lego, tang2023diffuscene}, or perform iterative 3D editing to add new compositional attributes \cite{cheng2023progressive3d, zhuang2023dreameditor, li2023focaldreamer}. The first track usually relies on a domain-specific dataset to learn scene priors, and the focus is different from ours on generating compositional 3D assets. While the second track can only add compistional attributes to the single target object through iterative editing, our methods can generate 3D assets with multiple compositional subjects or attributes in single round of training. 


\section{Preliminaries}
\subsection{Attend-and-Excite Revisited}
Attend-and-Excite \cite{chefer2023attend} was originally proposed to ease the catastrophic neglect issue in Text-to-Image generation domain, in which the text-guided image diffusion model can fail to generate one or more subjects specified in the target text prompt. 
Attend-and-Excite \cite{chefer2023attend} comes up a test-time optimization framework over the noisy latents, and encourages the cross-attention layers to attend to all subjects in the text during the iterative denoising process.\\

The intuition lies in the adopted cross-attention mechanism to bring in text condition into image generation. At each timestep $t$, the text embeddings will be fed into the cross-attention layer of the U-Net, and each latent feature over the feature grid will perform attention operation with all the text embeddings, resulting an attention activation matrix per text token. The attention matrix can be reshaped to obtain a spatial map $A_t^s \in \mathcal{R}^{H \times W}$ per text token $s$. Intuitively, for a token to be manifested in the generated image, there should be at least one patch in its map with a high activation value. To guide such desired behavior, Attend-and-Excite introduces $f(z_t) = \mathcal{L}_{att}=\underset{s\in \mathcal{S}}{\max} \, \mathcal{L}_s, \mathcal{L}_s = 1 - \text{max}({Gaussian}(A_t^s))$, where $z_t$ is the noisy latent. ${Gaussian}$ denotes applying Gaussian smoothing to the 2D activation map in order to cover a larger patch that later can emerge to the target objects. Such a loss will strengthen the activations of the most neglected subject token at the current timestep $t$.

\subsection{Multi-View Diffusion Models}
MVDream is a recent effort that adapts the common Text-to-Image diffusion model to have multi-view consistency, and enables Janus-free and high-fidelity Text-to-3D. 
Given a set of noisy images \(\mathbf{x}_t \in \mathrm{R}^{F\times H \times W \times C}\) , a text prompt as condition \(\mathbf{y}\), and a set of extrinsic camera parameters \(\mathbf{c} \in \mathrm{R}^{F\times 16}\), MVDream is trained to simultaneously denoise and generate multiple images  \(\mathbf{x}_0 \in \mathrm{R}^{F \times H \times W \times C}\)  that correspond to \(F\) different views of the same scene. At each step $t$, we have the predicted noise as  \(\epsilon_{\theta}(\mathbf{x}_t; \mathbf{y}, \mathbf{c}, t)\), where \(\theta\) denotes the parameters of the latent U-Net. To inherit the generalizability of the 2D diffusion models, while also obtaining the capability of multi-view consistency, MVDream fine-tunes on Stable Diffusion v2.1. However, MVDream also inherits the same issue as Stable Diffusion and can fail in generating compositionally correct 4-view images, and the lack of large-scale compositional scene-level 3D data for fine-tuning makes the issue more significant when applying to Text-to-3D. 



\section{Method}
\label{sec:method}
\begin{figure*}[ht]
\centering
\includegraphics[width=0.8 \linewidth]{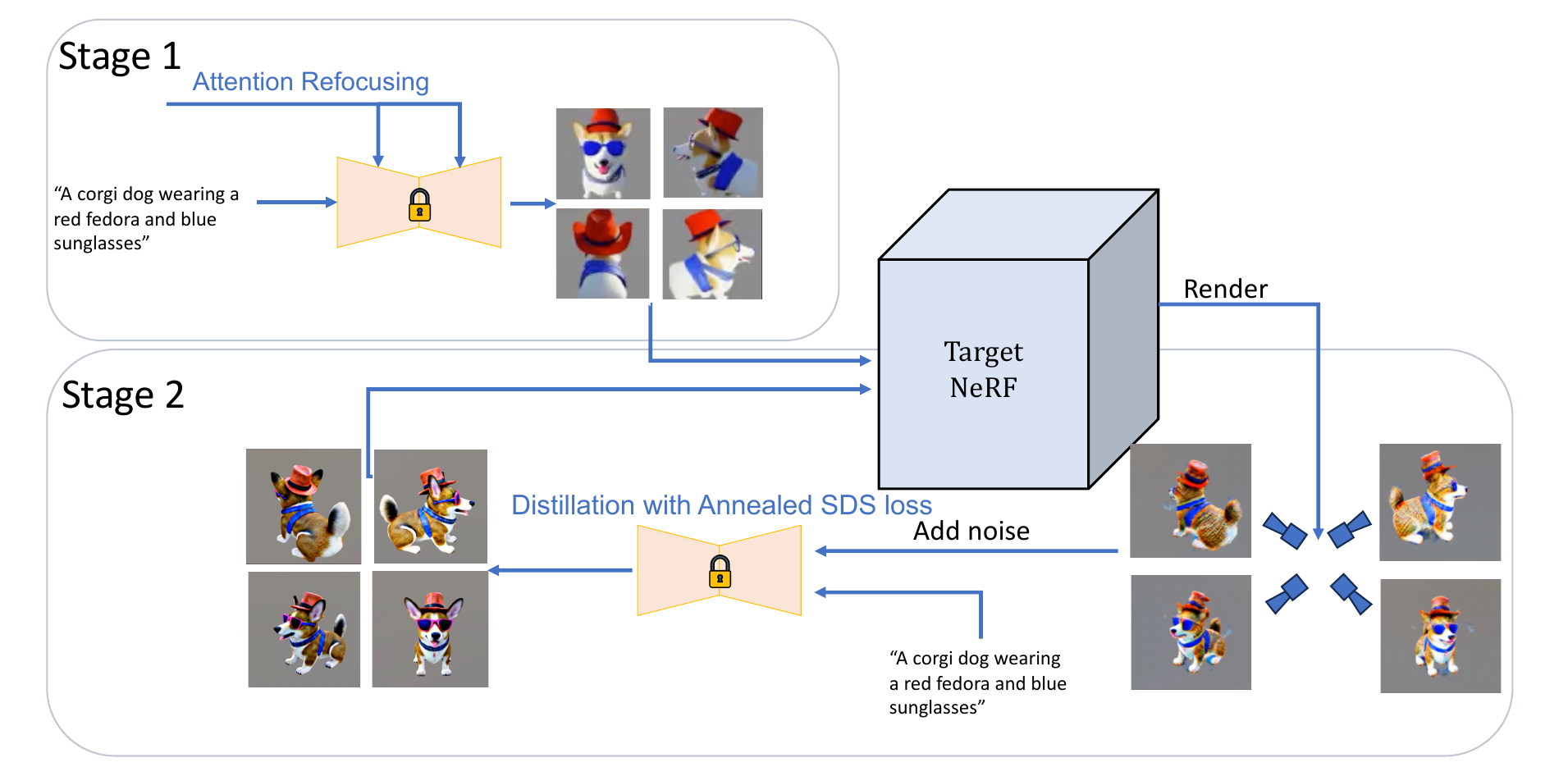}
\caption{\textit{Illustration of the two-stage pipeline with our Grounded-Dreamer.} Given a text prompt, we first generate compositionally correct 4-view images using iterative latent optimization at selected DDIM sampling steps. The 4-view reference images together with the masks are combined with score distillation sampling (SDS) loss in our hybrid training strategy, which will create high fidelity 3D assets while preserving the compositional priors accurately.}
\label{fig:pipe}
\end{figure*}
\vspace{2mm}
To tackle the specific challenges when generating compositionally correct 3D assets while achieving diversity, we propose a novel 2-stage approach that well incorporates attention refocusing mechanism and sparse-view guidance in a unified framework, as drawed in \cref{fig:pipe}.
In \cref{sec:stage1}, we detail our method for generating compositionally accurate four-view images. This stage focuses on ensuring that the subjects within these images are not only compositionally correct but also maintain the correct spatial relationships. Following this, in \cref{sec:stage2}, we explore the integration of these consistent reference images with a pre-trained multi-view diffusion model. Here, we examine the optimal combination of sparse-view reference images and Score Distillation Sampling (SDS) loss to achieve high-fidelity 3D asset generation.


\subsection{Attention Refocusing for Accurate Compositional 4-View Generation}
\label{sec:stage1}
Despite the success of Attend-and-Excite in Text-to-Image generation with multiple subjects, it is non-trivial extending the attention refocusing to Text-to-3D generation to optimize the target NeRF. 
Instead of optimizing a single-view latent, now we need to jointly optimize the 4-view latents without breaking the multi-view consistency, and we don't want to result in latent updates that lead to the latent becoming out-of-distribution. While it looks appealing to directly train a NeRF with combined attention refocusing loss and SDS loss, it renders a more challenging optimization problem since the attention refocusing loss is not optimizing the NeRF directly but on the rendered noisy latents from NeRF. The asynchronous NeRF updates can easily violate the assumption of in-distribution noisy latents on the attention refocusing loss. As we will show in the ablation study \cref{sec:1-stage}, such an attempt can easily lead to sub-optimal solutions and significantly enlarge the convergence time. \\
To design an more effective paradigm for composition control in Text-to-3D, we thus first adapt attention refocusing mechanism into compositionally correct 4-view generation, then we use the sparse-view images as additional compositional constraints in the 2nd-stage SDS-based NeRF training. When using the pre-trained multi-view diffusion model to generate 4-view images following a text prompt, we will have 
attention activation map $A_t^s \in \mathcal{R}^{F \times H \times W}$ per text token $s$, $F$ is the number of frames. Instead of naively updating the per-view latent using a per-view $\mathcal{L}_{att}$, we found that if first aggregating the attention maps across the 4 views using average operation, we tend to get more reasonable 4-view images, in which the final loss is 
\begin{equation}
\vspace{-2mm}
    \mathcal{L}_{att} = \underset{s\in \mathcal{S}}{\max}\, (1 - \max(\text{}\text{mean}({Gaussian}(A_t^s[v,:,:])))).
\end{equation}

Our final algorithm for applying attention refocusing in multi-view generation is drawn in Algorithm~\ref{alg:attend-and-excite}. Compared to Attend-and-Exicte, we also perform such optimization at more timesteps especially on the early stage instead of a few selected steps, which can still be done in minutes. 

\begin{algorithm}[h]
\caption{A Single Denoising Step on Compositional 4-View Generation}

\begin{algorithmic}[1] 
\label{alg:attend-and-excite}
\INPUT A text prompt $\mathbf{y}$, 4-view camera poses $\mathbf{c}$, a set of subject token indices $\mathcal{s}$, a timestep $t$, a set of iterations for refinement \{$t_1,\cdots,t_k$\}, a set of thresholds \{$T_1,\cdots,T_k$\}, and a trained multi-view diffusion model $SD_{mv}$.
\OUTPUT A noised latent $z_{t-1}$ for the next timestep
\begin{multicols}{2}
\STATE $\_, A_t \leftarrow SD_{mv}(z_t, \mathbf{y},\mathbf{c}, t)$ 
\STATE $A_t \leftarrow Softmax(A_t - \langle sot \rangle)$
\FOR{$s \in \mathcal{S}$}
    \STATE $A_t^s \leftarrow Mean_v(A_t[v,:,:,s])$
    \STATE $A_t^s \leftarrow Gaussian(A_t^s)$
    \STATE $\mathcal{L}_s \leftarrow 1 - max(A_t^s)$
\ENDFOR 
\STATE $\mathcal{L} \leftarrow max_s(\mathcal{L}_s)$
\STATE $z_t^\prime \leftarrow z_t - \alpha_t \cdot \Delta_{z_t}\mathcal{L}$
\IF{$t \in \{t_1,\cdots,t_k\}$}  
    \IF{$\mathcal{L} > 1-T_t$}
        \STATE $z_t \leftarrow z_t^\prime$ 
        \STATE \textbf{Go to} Step 1
    \ENDIF
\ENDIF
\STATE $z_{t-1}, \_ \leftarrow SD_{mv}(z_t^\prime, \mathbf{y},\mathbf{c}, t)$ 
\RETURN $z_{t-1}$
\end{multicols}
\end{algorithmic}

\end{algorithm}

\subsection{Coarse-to-Fine Synergistic Reconstruction With Diffusion Priors}
\label{sec:stage2}
We adopt an optimization-based reconstruction framework that leverages both the 4-view reference images, and a pre-trained multi-view diffusion model for priors-augmented reconstruction. To avoid running into the failure patterns as mentioned above, we hypothesize that the key lies in designing an effective training strategy that can create synergy between the two different supervision signals. Our key insight is that, the rough 4 reference views give coarse but nearly complete information about geometry of the target scene, especially on the compositional subjects, their interaction poses and spatial arrangement. These can provide a coarse initialization to later diffusion-based 3D distillation process. 

\mypara{Early few-shot NeRF training} At early stages, we hypothesize that GT reference images can be more informative and stable compared to SDS loss with the pre-trained MVDream diffusion model given a large timestep. Thus we introduce sparse-view NeRF to establish the coarse geometry and texture. We adopt the hierarchical hash-grid MLP introduced in Instant-NGP \cite{muller2022instant} as our learnable NeRF representation, and we can simply rely on RGB and mask reconstruction loss which are sufficient to obtain a coarse NeRF representation. The reconstruction loss is defined as $\mathcal{L}_{img} = \mathcal{L}_{RGB} + \mathcal{L}_{mask}$. 

\mypara{3D distillation with warm-start SDS loss} When the rough compositional geometry, and the associated texture emerge from the early few-shot NeRF training, we marry sparse-view NeRF with SDS-based 3D distillation to enable high-fidelity 3D generation, while preserving the compositional priors. The key idea is to bring in additional supervision from multiple unobserved viewpoints. In addition to the 4 fixed observed viewpoints that already have the roughly correct ground-truth images, we will randomly sample unobserved views and render multi-view images, the gradients come from SDS loss with a pre-trained multi-view diffusion model under text guidance. The total loss for NeRF training is defined as: 
\begin{align}
\mathcal{L}_{total} & = \lambda_{img} (i) * \mathcal{L}_{img}^{\text{fixed views}} + \lambda_{SDS} (i) * \mathcal{L}_{SDS}^{\text{random views}} \\
\mathcal{L}_{SDS}^{\text{random views}} & = \mathcal{E}_{t,\mathbf{c}, \epsilon} [\omega(t)\| \mathbf{\epsilon}_{\theta}(\mathbf{x}_t;y,\mathbf{c}, t(i)) -  \epsilon )\|_2^2]
\end{align}
When the NeRF representation of the target scene is optimized to a certain level, further relying on poor-quality 4-view reference images may hinder the creation of high-fidelity 3D assets. Hence, we choose to gradually reduce the weight of the image reconstruction loss to 0, while increasing the weight of the SDS loss. 



\begin{figure*}[ht]
\centering
\includegraphics[width=0.8 \linewidth]{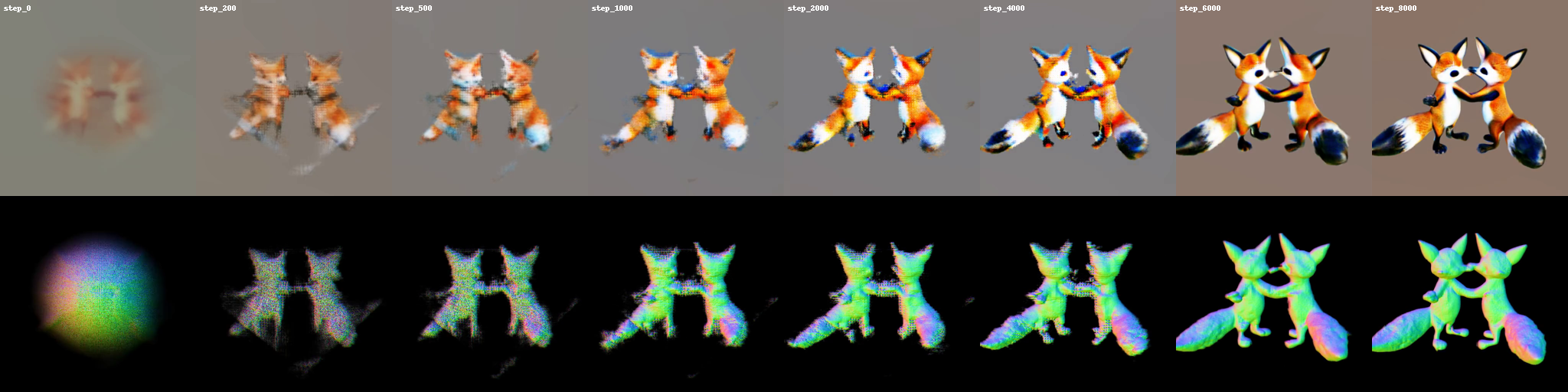}
\caption{\textit{Illustration on the 2nd-stage training progress with \METHODNAME.} Here we are showing a fixed front-view rendering of the target NeRF at different optimization steps. Our method can gradually create high fidelity 3D assets while preserving the compositional priors accurately.}
\label{fig:progress}
\end{figure*}

We have $t(i) \sim \mathcal{U}(T_{min}(i), T_{max}(i))$.
In prior works \cite{zhu2023hifa, huang2023dreamtime, shi2023mvdream} a time-annealing approach is implemented, where $T_{max}(i_{start})$ is set close to the total number of timesteps. We found such implementation can lead to large variation in SDS loss and drastic content change in NeRF output towards entirely different directions. It can raise the resurfacing alignment issues with the compositional priors drawn from the reference images. To solve this and preserve composition, the key modification we made is to set the initial \(T_{max}(i_{start}) \) as small as 680, thus the SDS loss can be leveraged to add more details, and refine NeRF to high fidelity. Specifically, 
\begin{align}
    T_{max}(i) & = c_1 + (c_2-c_1) * \frac{i-a}{b-a}
\end{align}
The above hierarchical training design and modification to timestep annealing are simple but critical. It works effectively to have SDS loss and sparse image supervisions to make synergy between each other, an example is shown in \cref{fig:progress}. We will next demonstrate through experiments and ablations.

\section{Experiments}
\label{sec:exp}

\subsection{Experimental Setup}

\mypara{Implementation details} We use the same settings for our method on all the text prompts. The early few-shot NeRF is trained for 200 steps, and then we add SDS loss by setting the \(T_{max}(i) \) to linearly reduce from 680 to 500, while \(T_{min} (i) \) is linearly reduced from 380 to 20. During the first 5000 steps we train the NeRF at 64x64 resolution, and switch to 256x256 for the latter 5000 steps for refinement. We gradually reduce the weighting on image reconstruction loss from 1000 to 100, while the weighting on SDS loss is increased from 0.025 to 0.25. For most of the baselines, we adopt default implementations within \cite{threestudio2023}, and we run Wonder3D \cite{long2023wonder3d} using their released code, and use the scripts provided by Magic123 \cite{qian2023magic123} for background removal. All the experiments are conducted on Nvidia A100 GPUs. 

\mypara{Prompts set} To cover various scenarios of compositional Text-to-3D, we select and categorize our text prompts into (1) compositional-objects, which involve multiple subjects arranged in specific spatial relationships, e.g. ``a green spoon on a red cake in a yellow tray''; (2) compositional-animals, which contain scenarios of animal-object interaction, or specific activities, e.g. ``An artist is painting on a blank canvas'' or ``two foxes fighting''. We select 50 for each subgroup based on existing text prompts from \cite{poole2023dreamfusion}\ and \cite{cheng2023progressive3d}, with 100 text prompts in total. 

\mypara{Baselines} 
We consider MVDream~\cite{shi2023mvdream} as our baseline for multi-view generation. 
For text to 3D generation, we adopt recent SOTA Text-to-3D methods such as Magic3D \cite{lin2023magic3d}, ProlificDreamer \cite{wang2023prolificdreamer}, and the MVDream-ThreeStudio \cite{shi2023mvdream} as the baselines. To further illustrate the unique advantages of 4-view input, we compare with recent single-view-to-3D methods like Magic-123 \cite{qian2023magic123} by first using the text to generate a proper single-view image. To demonstrate the unique benefits of our hierarchical NeRF optimization with text guided diffusion priors, we also compare the results with a recent SOTA image-conditioned multi-view diffusion model named Wonder3D \cite{long2023wonder3d}, which is essentially a single-view-to-4-view-to-3D method. Wonder3D learns 3D priors that are capable of generating normal maps and RGB images on 3 additional viewpoints, however, they adopt a normal-assisted few-shot NeRF to get the final 3D assets.

\mypara{Metrics} 
CLIP-based metrics might fail to measure the fine-grained correspondences between described
objects and binding attributes, thus we only use CLIP R-Precision following \cite{poole2023dreamfusion}, which measures the relative closeness between all generated images and their corresponding text prompts. Following a recent effort \( T^{3} \text{Bench} \) \cite{he2023t}, we adopt a VQA-like pipeline by first using image-to-text models like BLIP-2 \cite{li2023blip} to generate captions for rendered multi-view images, and then evaluate the alignment score between the compressed predicted captions and the given text prompt, named as T3 Score II. The purpose is to evaluate the capability of handling general text prompts with complex semantics in a fine-grained manner. We also conduct user study and add human preferences score on text-image alignment as additional metric. For measuring images realism, we compute the FID score following \cite{heusel2017gans}. 

\subsection{Enhanced Compositional 4-view Synthesis}
\begin{figure*}[!ht]
    \centering
    \includegraphics[width=0.8 \linewidth]{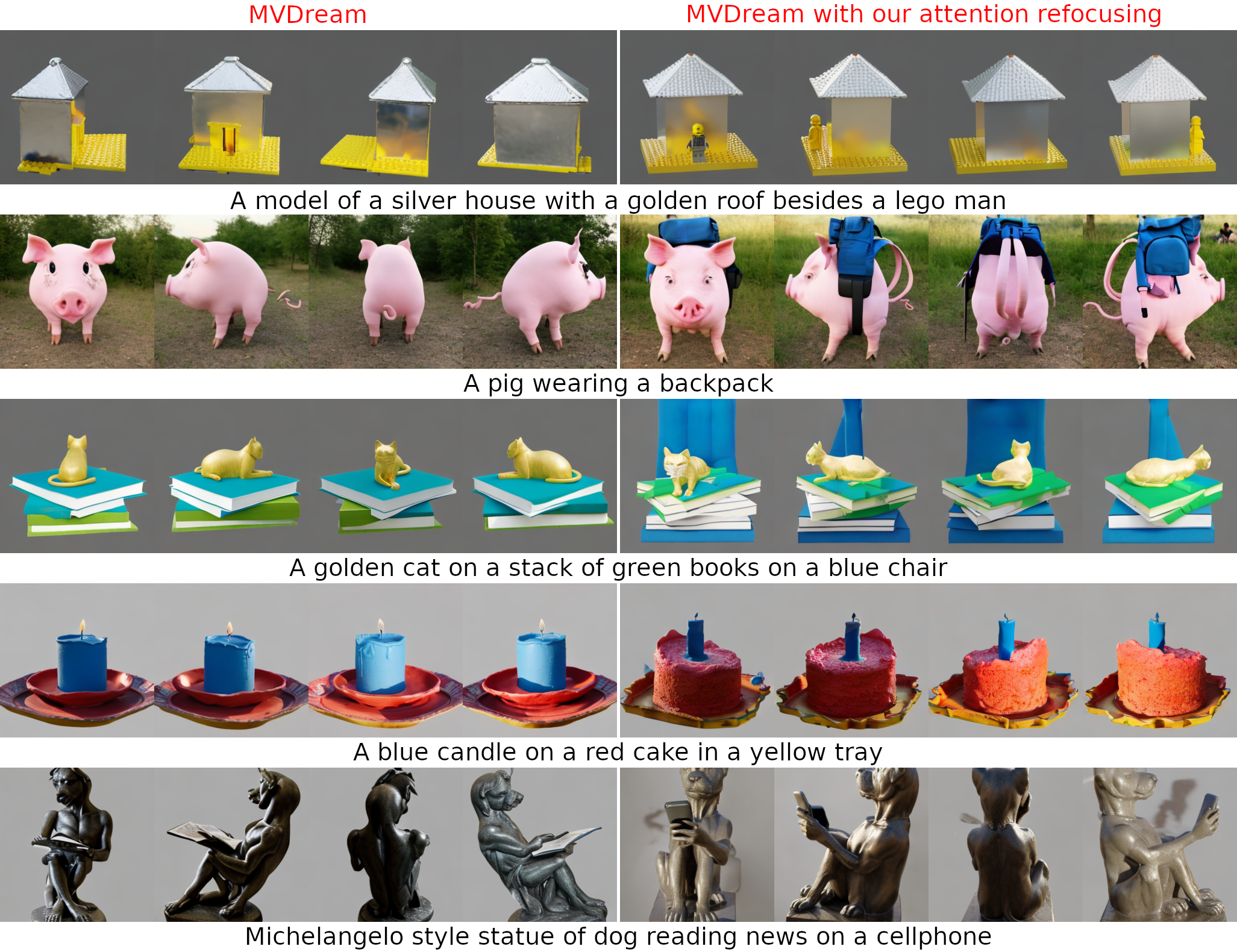}
    \captionof{figure}{\textit{4-view generation, each pair uses the same random seed}. Our inference-stage optimization encourages compositionally correct 4-view generation compared to the original MVDream.} 
    \label{fig:t4v_compare}
\end{figure*}

We show both quantitative and qualitative results in \cref{fig:t4v_compare} and \cref{tab:t4v_compare}. 
 Our inference-stage editing method improves on the text-image alignment, while achieving comparable image realism. 

\begin{table}[h!]
  \centering
  \scriptsize
  \begin{tabular}{l|c|c}
    \hline
    Method & FID Score $\downarrow$ & T3 Score II (\%) $\uparrow$\\
    \hline\hline
    \texttt{MVDream} & \first{71.43} & 2.72 \\
    \texttt{Ours} & 74.16 & \first{2.85} \\
    \hline
  \end{tabular}
  \caption{\textit{Quantitative evaluation on 4-view generation}. We run each method on 100 prompts with different random seeds, and compare the CLIP score of the generated images. Our inference-stage editing can generate more accurate images regarding the composition of different target subjects, while not breaking the multi-view consistency.}
  \label{tab:t4v_compare}
  \vspace{-5mm}
\end{table}

\subsection{Compositional and Diverse 3D Generation}
\begin{figure*}[h]
    \centering
    \includegraphics[width=1.0 \linewidth]{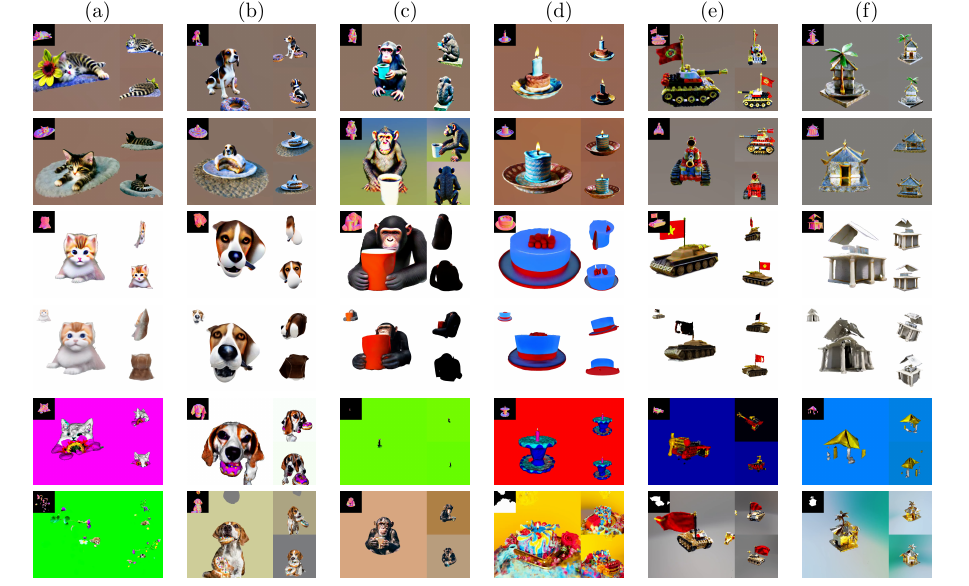}
    \caption{\textit{Qualitative results comparison for compositional Text-to-3D}. From top to down, the methods are: our Grounded-Dreamer, MVDream \cite{shi2023mvdream}, Magic123 \cite{qian2023magic123}, Wonder3D \cite{long2023wonder3d}, Magic3D \cite{lin2023magic3d}, ProlificDreamer \cite{wang2023prolificdreamer}. Our method generates more compositionally complete views with high quality. Text prompts: (a) ``a zoomed out DSLR photo of an adorable kitten lying next to a flower'', 
    (b) ``a zoomed out DSLR photo of a beagle eating a donut'',
    (c) ``a zoomed out DSLR photo of a chimpanzee holding a cup of hot coffee'',
    (d) ``a blue candle on a red cake in a yellow tray'',
    (e) ``a lego tank with a golden gun and a red flying flag'',
    (f) ``a model of a silver house with a golden roof beside an origami coconut tree''.}
    \label{fig:t3d_compare}
    \vspace{-5mm}
\end{figure*}

We show quantitative and qualitative comparisons in \cref{tab:t3d_compare} and \cref{fig:t3d_compare} for text-guided 3D generation with multiple subjects  present. As detailed in \cref{tab:t3d_compare}, our method consistently outperforms existing SOTA baselines considering the overall performance of text-image alignment, view quality and consistency. 

Specifically, our approach excels in the CLIP-R-Precision and T3 Score II metrics, indicative of superior performance in consistent text-guided 3D generation.  In terms of view consistency, our method performs on par with MVDream while significantly exceeding all other models.
Compared with MVDream, our method largely outperforms it for compositional generation. As illustrated in \cref{fig:t3d_compare}, MVDream frequently overlooks certain compositional elements, leading to incomplete or imprecise representations. In contrast, our method generates more compositionally complete views. Also thanks to the hierarchical optimization strategy, our method achieves high-fidelity 3D generation as evidenced by the FID score. ProlificDreamer, on the other hand, also achieves remarkable visual quality, producing sharp and highly detailed results, however it suffers from slow training speeds (refer to our supplementary material for an efficiency report) and is prone to issues such as corrupted flat geometries or `Janus' problems, as shown in the case of row 4. From \cref{tab:t3d_compare} we can also see its inferior view consistency performance.

\setlength{\tabcolsep}{1pt}

\begin{table}[t]
  \centering
  \scriptsize
  \resizebox{\linewidth}{!}{\begin{tabular}{l|c|c|c|c|c}
    \hline
    Method & T3 Score II $\uparrow$ & CLIP R-P.(\%) $\uparrow$ & Good alignment $\uparrow$ & Freq. of Janus$\downarrow$ &   FID Score $\downarrow$  \\
    \hline\hline
    \texttt{Magic3D} & 2.29/5.0 & 27.10 & 27.10 & 58.88 & 137.45 \\
    \texttt{ProlificDreamer} & \first{2.68/5.0} & \second{48.91}  & \first{60.87} & 77.17 & 129.11 \\
    \texttt{Magic-123} & 2.32/5.0 & 24.74 & 28.87 & 64.95 & 121.16 \\
    \texttt{Wonder3D} & 1.96/5.0 &  20.22 & 35.60 & 21.25 & 129.93  \\
    \texttt{MVDream} & 2.33/5.0 & 44.95 & 44.44 & \first{5.88} & \first{109.78} \\
    \texttt{Ours} & \second{2.53/5.0} & \first{62.73} & \second{56.71} & \second{17.15} &\second{115.94}   \\
    \hline
  \end{tabular}
  }
  \caption{\textit{Quantitative evaluation on 3D composition}. We run each method on 100 prompts with same random seeds. Under ``Good alignment”, for each method, we show the percentage of the generated 3D outputs that human reviewers annotate as aligning well with the text prompts.  Under ``Freq. of Janus'', we show the preference ratio of generated outputs for each model in terms of view consistency. }
  \label{tab:t3d_compare}
\end{table}
Our method also largely outperforms single-image-to-3D approaches like Magic-123 \cite{qian2023magic123}, Wonder3D \cite{long2023wonder3d} on the benchmark text prompts. While Magic-123 can effectively reconstruct the front view of objects, it struggles with side or back perspectives, leading to incorrect anatomy, like in the case ``a zoomed out DSLR photo of a chimpanzee
holding a cup of hot coffee". Other recent works, such as Wonder3D, typically face challenges in reconstruction quality, particularly when relying solely on sparse-view images. In contrast, our method innovatively integrates text-guided natural image priors with sparse-view supervision. Our generated 3D assets can harvest the natural images priors from a pre-trained diffusion model, which not only enhances the quality of reconstruction but also ensures generalizability across various text prompts describing a wide range of 3D compositional subjects. 
The results demonstrate the capability of our method in achieving high-quality, compositionally accurate 3D reconstructions without compromising details and structural integrity.

Compared to ProlificDreamer, our method can generate diverse 3D assets in a well time-bounded manner, as shown in \cref{fig:t3d_diverse}. The diversity can be easily controlled with different pairs of edited 4-view images as additional guidance. 

\section{Ablation Study}
\begin{figure}[h]
    \centering
    \includegraphics[width=1.0\linewidth]{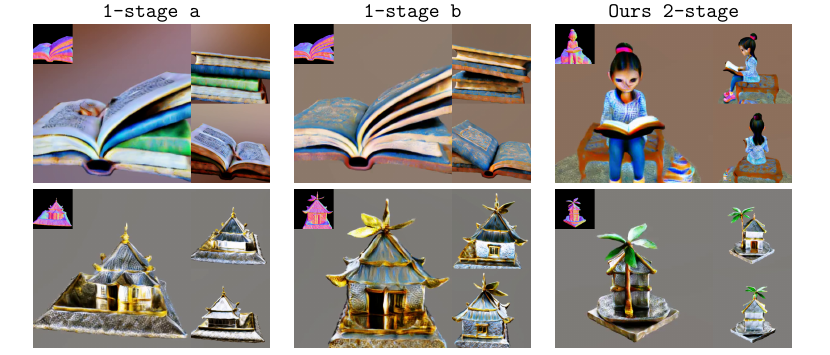}
    \caption{\textit{Results visualization with different pipeline designs}.
}
    \label{fig:number_stage}
\label{fig:figure1}
\end{figure}

\mypara{1-stage design with attention refocusing loss}
\label{sec:1-stage}
We also implement and train two 1-stage variants without first generating 4-view images, each with balanced losses: a) we directly add the attention score loss to the SDS loss; b) we first update latents using the attention loss, then we use the updated noisy latents for the SDS loss. \cref{tab:number_stage} show the ablation results. 
Compared to our 2-stage approach: 1) the 1-stage ones significantly increase the training time by two to three times; 2) stark performance drop on compositional accuracy; 3) corrupted shapes, like the tree mixes with the house in the 2nd row of \cref{fig:number_stage}. 

\begin{figure}
    \centering
    \begin{tabular}{cc}
       \texttt{Ours-naive} \\
         \includegraphics[width=0.8\linewidth] {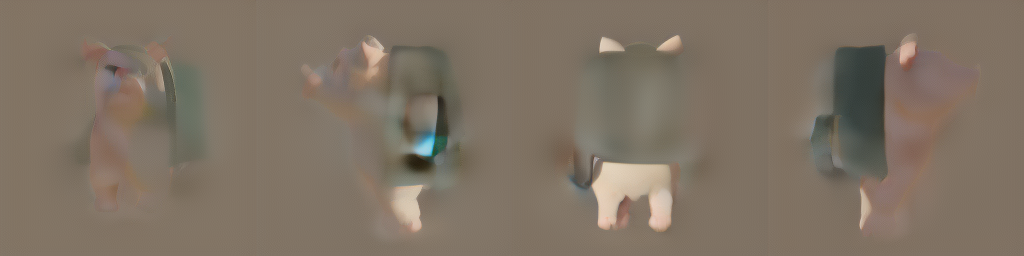} \\
         \texttt{Ours-final}  \\
        \includegraphics[width=0.8\linewidth]{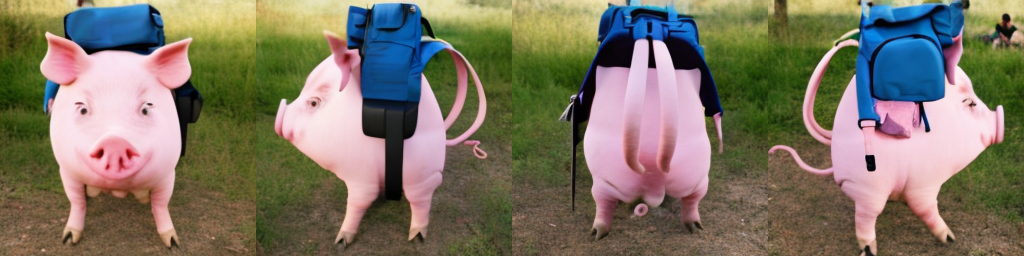} 
    \end{tabular}
    \caption{\textit{Examples on ``a pig wearing a backpack''.}}
\label{fig:sequential}
\end{figure}

\begin{table}[t]
  \centering
  \scriptsize
  \begin{tabular}{l|c|c|c}
    \hline
    Method & FID Score $\downarrow$ & CLIP R-P. (\%) $\uparrow$ & GPU-hours $\downarrow$ \\
    \hline\hline
    \texttt{Ours-2-stage} & 115.94 & \first{62.73} & \first{1.91} \\
    \texttt{Ours-1-stage a} & 114.68 & 48.18 & 3.90 \\
    \texttt{Ours-1-stage b} & \first{112.45} & 40.91 & 5.25 \\
    \hline
  \end{tabular}
  \caption{\textit{Ablation with 1-stage designs}.}
  \label{tab:number_stage}
  \vspace{-.5cm}
\end{table}

\mypara{Different strategies for multi-view latent updates}
For example, if naively updating each view sequentially using Attend-and-Excite, we can end up with corrupted views as shown in \cref{fig:sequential}. Our later mean operation across the 4 views latents is simple and already works sufficiently well on creating compositionally correct 4-view images.
\setlength{\tabcolsep}{1pt}

\begin{figure}[h]
    \centering
    \includegraphics[width=1.0\linewidth]{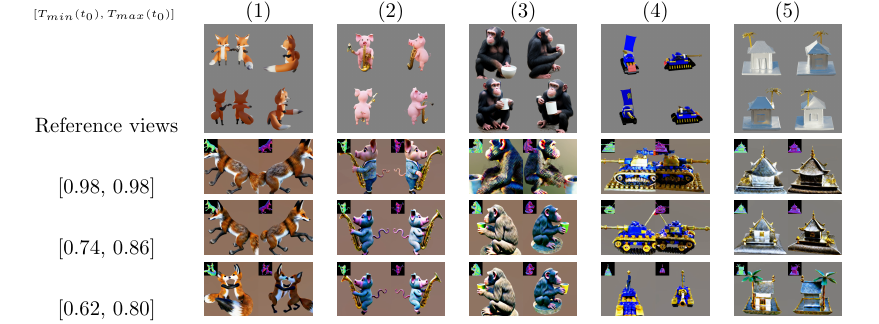}
    \caption{\textit{3D generation under different initial timestep sampling range.} We pick 5 text prompts, and visualize the reference 4-view images, and two side-view of the generated 3D assets with associated normal images attached to the top-left corner.}
    \label{fig:t3d_timestep}
\end{figure}

\mypara{Effects of warm-start timestep in SDS loss}
One of the keys to our method's success is to set the initial timestep of SDS loss to be relatively smaller, so the SDS loss can guide NeRF towards adding more geometric and texture details, while preserving the compositional priors. As shown in \cref{fig:t3d_timestep}, if following the default setting like \( [T_{min}(t_0), T_{max}(t_0)] = [0.98, 0.98] \) out of 1000, or\( [T_{min}(t_0), T_{max}(t_0)] = [0.74, 0.86] \), it can miss the compositional information like ``two'' in the first example, or ``beside an origami coconut tree'' in the 5th example. 

\mypara{Effects of backbone used in SDS loss}
To validate the effects of using pre-trained multi-view diffusion model, we can replace our Score Distillation Sampling (SDS) loss backbone with SD v2.1. The results, as depicted in the accompanying \cref{tab:t3d_ablation_sd}, demonstrate that using pre-trained multi-view diffusion model for SDS loss help consistently yield better textural quality.

\begin{table}[h!]
  \centering
  \scriptsize
  \begin{tabular}{l|c|c}
    \hline
    Method  & Freq. of Janus (\%) $\downarrow$ & FID Score $\downarrow$ \\
    \hline \hline
    \texttt{ours with SD} & 40.90 & 135.60 \\
    \texttt{Ours} & \first{17.15} & \first{115.94} \\
    \hline
  \end{tabular}
  \caption{\textit{Text-to-3D with different pretrained T2I models}. }
  \label{tab:t3d_ablation_sd}
\end{table}

\section{Conclusion}
In summary, our work introduces a novel two-stage framework for Text-to-3D synthesis, effectively overcoming challenges in compositional accuracy and diversity. The first stage leverages a multi-view diffusion model for generating spatially coherent views from text, while the second stage synergizes sparse-view NeRF with text-guided diffusion priors for refined 3D reconstruction. This approach not only enhances the fidelity and compositional integrity of 3D models from complex text prompts, but also paves the way for future explorations in seamless 2D-to-3D transitions and model versatility. Our method demonstrates a significant leap in Text-to-3D synthesis, offering a robust foundation for further advancements in this evolving field.

\clearpage
{
    \small
    \bibliographystyle{ieeenat_fullname}
    \bibliography{main}
}

\setcounter{page}{1}
\maketitlesupplementary

\begin{figure*}[ht]
    \centering
    \includegraphics[width=1.0 \linewidth]{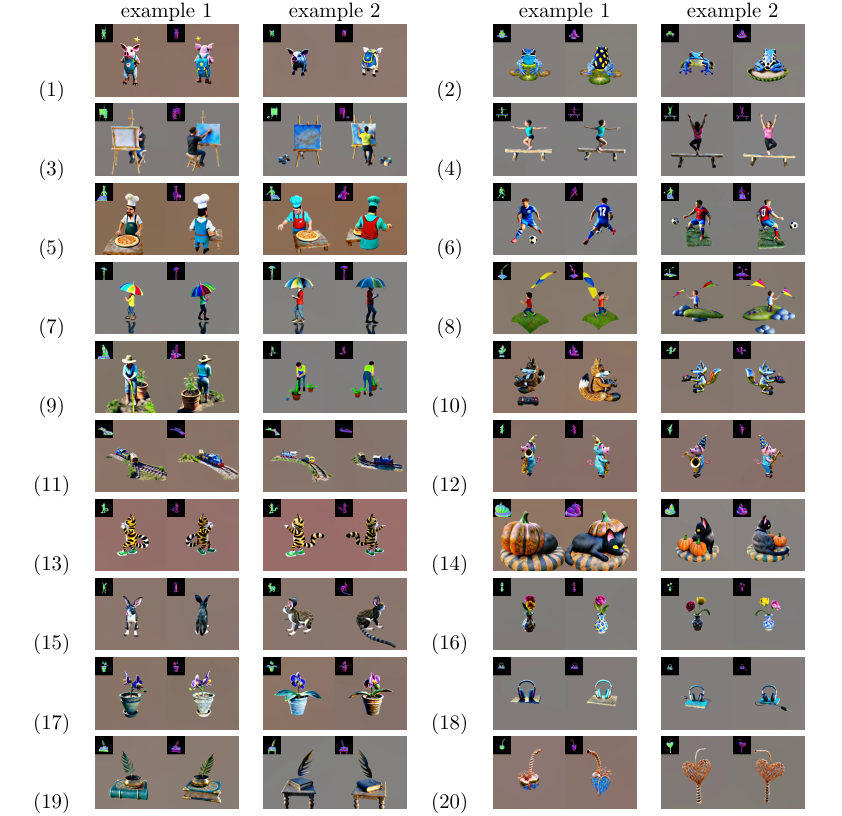}
    \caption{\textit{Diverse 3D assets generated with \METHODNAME}. We show results for 20 text prompts in total, and for each text prompt, we visualize two different generated samples, using their two side-view with associated normal images attached to the top-left corner. (1) "a pig wearing a backpack",
(2) ``a blue poison-dart frog sitting on a water lily",
(3) ``An artist is painting on a blank canvas",
(4) ``A young gymnast trains with a balance beam",
(5) ``A chef is making pizza dough in the kitchen",
(6) ``A footballer is kicking a soccer ball",
(7) ``A man is holding an umbrella against rain",
(8) ``A boy is flying a colorful kite in the sky",
(9) ``A gardener is watering plants with a hose",
(10) ``a DSLR photo of a fox holding a videogame controller",
(11) ``a DSLR photo of Two locomotives playing tug of war",
(12) ``a zoomed out DSLR photo of a pig playing the saxophone",
(13) ``an orange cat wearing a yellow suit and green sneakers and pink cap",
(14) ``A black cat sleeps peacefully beside a carved pumpkin",
(15) ``a monkey-rabbit hybrid",
(16) ``a blue peony and a yellow tulip in a pink vase",
(17) ``an orchid flower planted in a clay pot",
(18) ``a DSLR photo of a pair of headphones sitting on a desk",
(19) ``a DSLR photo of a quill and ink sitting on a desk",
(20) ``a DSLR photo of A very beautiful tiny human heart organic sculpture made of copper wire and threaded pipes, very intricate, curved, Studio lighting, high resolution".}
\label{fig:t3d_additional}
\end{figure*}

In this supplementary material, we provide additional implementation details in \cref{sec:add_implement}, offering deeper insights into our methodology. For a comprehensive understanding of our method's efficiency, a comparative speed analysis against selected methods is detailed in \cref{sec:speed}. We also discuss the failure modes of our approach in \cref{sec:failure}, highlighting areas for potential improvement. Additionally, \cref{sec:additional_viz} includes an expanded set of qualitative visualizations, complementing those presented in \cref{fig:t3d_additional}, to better illustrate the capabilities and limitations of our method. Finally, you can find all text prompts we used for our experiments in \cref{sec:prompts_list}.

\section{Additional Implementation Details}
\label{sec:add_implement}
During stage 1, we use the pre-trained multi-view text-to-image (T2I) diffusion model from \cite{shi2023mvdream} to perform inference-stage editing and generate 4 views, we denote this model as MVDream-T2I. Our offline editing approach treats the generation of 4-view, text-aligned images as a latent optimization problem. In this framework, the attention score map associated with each subject token forms the basis of our objective function. This function is then used to optimize the latent noise at selected steps. During the 50-step DDIM-based inference sampling, we specifically target iterations 0,1,2,3,4,5,10,20,30,40 for latent optimization. This process is capped at a maximum of 25 iterations, with a loss threshold set at 0.1. This additional optimization step does extend the inference time from an initial 10 seconds to approximately 45 seconds. In stage 2, we also use the pre-trained MVDream-T2I model for SDS loss, without doing any fine-tuning or re-training.

\paragraph{Metrics}
In our main paper Tab. 2, we adopt ``Good alignment" and ``Freq. of Janus", both are new metrics proposed and defined in another work that we are submitting. Basically ``Good alignment" shows the percentage of the generated 3D models that that human reviewers annotate as aligning well with the text prompts, while ``Freq. of Janus" measures how often the Janus problem appears in the generated 3D content. 

\section{Speed Comparison}
\label{sec:speed}
While our primary focus is not on efficiency, our method is capable of generating high-quality 3D assets within a reasonable time of approximately 1.9 hours. In contrast, while ProlificDreamer \cite{wang2023prolificdreamer} also achieves diverse results, it does so at the cost of time efficiency, requiring upwards of 15 hours to produce final 3D assets. Our method strikes a more balanced approach, offering both diversity and a more manageable time investment. We anticipate that future developments in this field will further enhance the efficiency of our method. 
\begin{table}[hbt]
  \centering
  \begin{tabular}{|l|c|}
    \hline
    Method &  GPU-hours $\downarrow$\\
    \hline\hline
    \rowcolor{lightgray}\texttt{Magic3D}~\cite{lin2023magic3d}                       &  2.40   \\
    \rowcolor{lightgray}\texttt{ProlificDreamer}~\cite{wang2023prolificdreamer}      &  15.50  \\
    \rowcolor{lightgray}\texttt{MVDream}~\cite{shi2023mvdream}                       &  1.83  \\
    \texttt{Ours -- 1st stage}                                   &  0.02  \\
    \texttt{Ours -- full model}                                  &  1.91  \\
    \hline
  \end{tabular}
  \caption{\textit{GPU-hours needed to generate one 3D model given a prompt}.}
  \label{tab-efficiency}
  \vspace{-0.1in}
\end{table}

\section{Failure Cases}
\label{sec:failure}
\begin{figure*}[h]
    \centering
    \begin{tabular}{ccccccr}
        failure type & reference views  & 3D generation & failure type & reference views  & 3D generation &  \\
        (A)  &
        \includegraphics[width=0.15\linewidth]{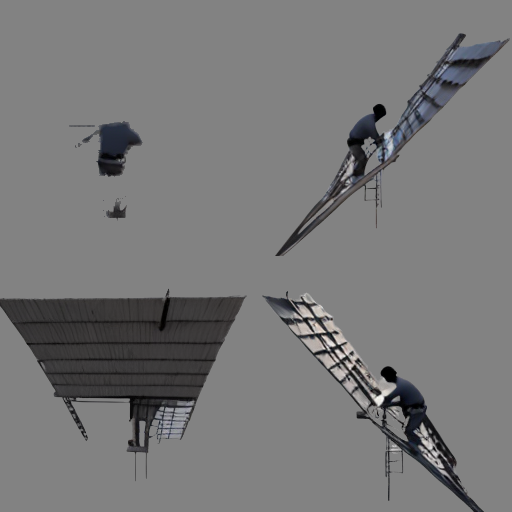} & 
        \includegraphics[width=0.15\linewidth]{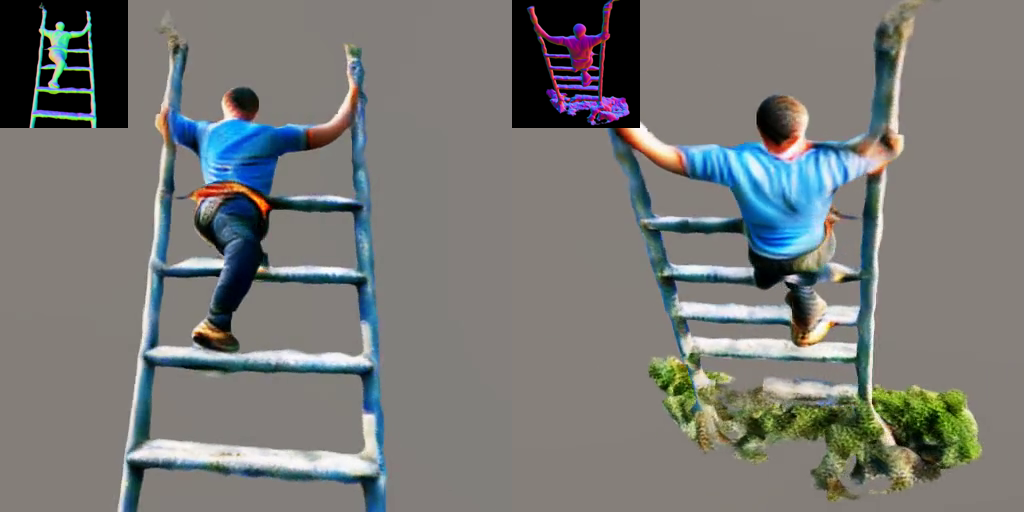} & 
        (B)  &
        \includegraphics[width=0.15\linewidth]{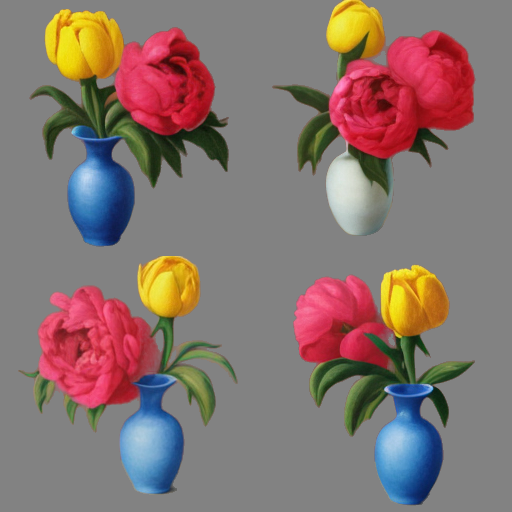} & 
        \includegraphics[width=0.15\linewidth]{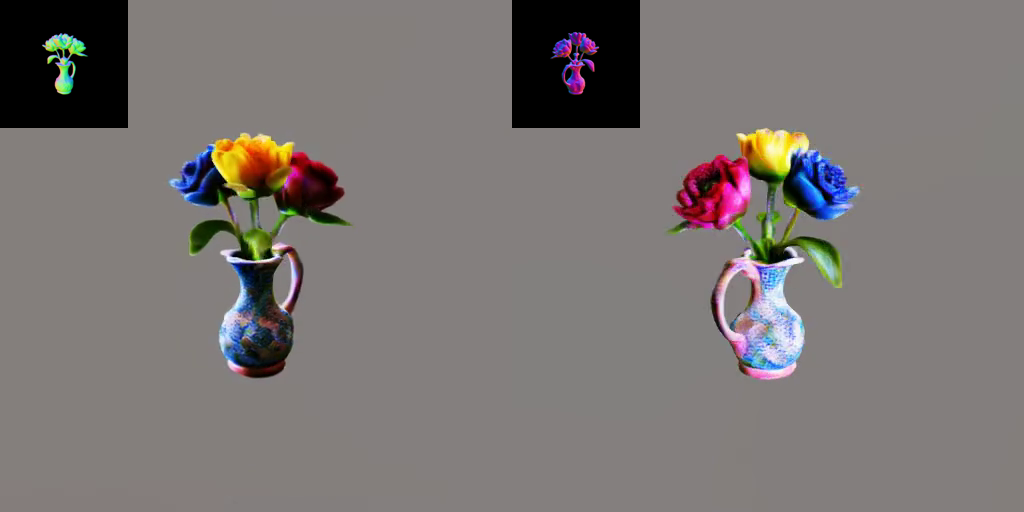} & \\
    \end{tabular}
    \caption{\textit{Failure cases with \METHODNAME}. We visualize the two side-view of the generated 3D assets with associated normal images attached to the top-left corner. Text prompts: 
 (1) ``A worker is climbing a ladder to repair a roof"; (2)
   ``a blue peony and a red rose and a yellow tulip in a pink vase". }
    \label{fig:failure}
\end{figure*}

Our analysis identifies two primary failure modes. The first involves incomplete foreground segmentation, leading to the omission of key compositional elements. For instance, as depicted in \cref{fig:failure}, when the roof of a house is not included in three views of the segmented final images, our Stage 2 process is unable to fully compensate for these missing elements. We expect more advanced segmentation tool like SAM \cite{kirillov2023segany} to help mitigate such issue. The second failure pattern relates to inaccuracies in additional attributes, particularly on color attribute in our case. An example of this, shown in \cref{fig:failure}, is the inability of both the initial four-view generation and the final 3D asset to accurately represent specified colors, such as yellow and blue. These limitations predominantly stem from the current capabilities and precision of the diffusion-based Text-to-Image (T2I) model backbone. 

\section{List of Prompts}
\label{sec:prompts_list}
In our actual implementation, we end up with having 111 text prompts, with the first 55 for capturing interactive compositional animals, and the last 56 involving static compositional objects with specific spatial arrangement. 

\small
\begin{enumerate}
 \item  ``a pig wearing a backpack" 
 \item  ``a blue poison-dart frog sitting on a water lily" 
 \item  ``a bumblebee sitting on a pink flower" 
 \item  ``a crocodile playing a drum set" 
 \item  ``a corgi dog wearing a red fedora and blue sunglasses" 
 \item  ``two foxes fighting" 
 \item  ``Michelangelo style statue of dog reading news on a cellphone" 
 \item  ``a tiger playing the violin" 
 \item  ``An artist is painting on a blank canvas" 
 \item  ``A student is typing on his laptop" 
 \item  ``A young gymnast trains with a balance beam" 
 \item  ``A chef is making pizza dough in the kitchen" 
 \item  ``A footballer is kicking a soccer ball" 
 \item  ``A man is holding an umbrella against rain" 
 \item  ``A girl is reading a hardcover book in her room" 
 \item  ``A woman putting lipstick on in front of a mirror" 
 \item  ``A worker is climbing a ladder to repair a roof" 
 \item  ``A florist is making a bouquet with fresh flowers" 
 \item  ``A boy is flying a colorful kite in the sky" 
 \item  ``A gardener is watering plants with a hose" 
 \item  ``A photographer is capturing a beautiful butterfly with his camera" 
 \item  ``A scientist is examining a specimen under a microscope" 
 \item  ``A drummer is beating the drumsticks on a drum" 
 \item  ``A fisherman is throwing the fishing rod in the sea" 
 \item  ``A baby is reaching for a teddy bear on the bed" 
 \item  ``a DSLR photo of a corgi wearing a beret and holding a baguette, standing up on two hind legs" 
 \item  ``a DSLR photo of a fox holding a videogame controller" 
 \item  ``a DSLR photo of a ghost eating a hamburger" 
 \item  ``a DSLR photo of a humanoid robot using a laptop" 
 \item  ``a DSLR photo of a koala wearing a party hat and blowing out birthday candles on a cake" 
 \item  ``a DSLR photo of Two locomotives playing tug of war" 
 \item  ``a lego man wearing a chef\u2019s hat and riding a golden motorcycle" 
 \item  ``a lego man wearing a silver crown and riding a golden motorcycle" 
 \item  ``a lego man wearing a wooden top hat and riding a golden motorcycle" 
 \item  ``a metal monkey wearing a chef\u2019s hat and driving an origami sport car" 
 \item  ``a metal monkey wearing a golden crown and driving an origami sport car" 
 \item  ``a metal monkey wearing a wooden top hat and driving an origami sport car" 
 \item  ``a zoomed out DSLR photo of a colorful camping tent in a patch of grass" 
 \item  ``a zoomed out DSLR photo of a dachsund riding a unicycle" 
 \item  ``a zoomed out DSLR photo of a hippo biting through a watermelon" 
 \item  ``a zoomed out DSLR photo of a monkey riding a bike" 
 \item  ``a zoomed out DSLR photo of a pig playing the saxophone" 
 \item  ``a zoomed out DSLR photo of a raccoon astronaut holding his helmet" 
 \item  ``a zoomed out DSLR photo of a tiger eating an ice cream cone" 
 \item  ``a zoomed out DSLR photo of an adorable kitten lying next to a flower" 
 \item  ``a wide angle zoomed out DSLR photo of A red dragon dressed in a tuxedo and playing chess. The chess pieces are fashioned after robots" 
 \item  ``a zoomed out DSLR photo of a beagle eating a donut" 
 \item  ``a zoomed out DSLR photo of a chimpanzee holding a cup of hot coffee" 
 \item  ``an orange cat wearing a yellow suit and cyan boots" 
 \item  ``an orange cat wearing a yellow suit and green sneakers" 
 \item  ``an orange cat wearing a yellow suit and green sneakers and pink cap" 
 \item  ``an orange cat wearing a yellow suit and red pumps" 
 \item  ``a golden cat on a stack of green books on a blue chair" 
 \item  ``A black cat sleeps peacefully beside a carved pumpkin" 
 \item  ``a monkey-rabbit hybrid" 
 \item  ``a blue candle on a red cake in a yellow tray" 
 \item  ``a blue peony and a red rose and a yellow tulip in a pink vase" 
 \item  ``a blue peony and a yellow tulip in a pink vase" 
 \item  ``a bunch of colorful marbles spilling out of a red velvet bag" 
 \item  ``a ceramic tea pot and a cardboard box on a golden table" 
 \item  ``a ceramic tea pot and a lego car on a golden table" 
 \item  ``a ceramic tea pot and a pair of wooden shoes on a golden table" 
 \item  ``a ceramic tea pot and an origami box and a green apple on a golden table" 
 \item  ``a ceramic tea pot and an origami box on a golden table" 
 \item  ``a ceramic upside down yellow octopus holding a blue green ceramic cup" 
 \item  ``a purple rose and a red rose and a yellow tulip in a pink vase" 
 \item  ``a red rose in a hexagonal cup on a star-shaped tray" 
 \item  ``a white tulip and a red rose and a yellow tulip in a pink vase" 
 \item  ``a zoomed out DSLR photo of a beautifully carved wooden knight chess piece" 
 \item  ``an orchid flower planted in a clay pot" 
 \item  ``a nest with a few white eggs and one golden egg" 
 \item  ``a pink peach in a hexagonal cup on a round cabinet" 
 \item  ``a plate of delicious tacos" 
 \item  ``a DSLR photo of a bagel filled with cream cheese and lox" 
 \item  ``a DSLR photo of a beautiful violin sitting flat on a table" 
 \item  ``a DSLR photo of a candelabra with many candles on a red velvet tablecloth" 
 \item  ``a DSLR photo of a Christmas tree with donuts as decorations" 
 \item  ``a DSLR photo of a cup full of pens and pencils" 
 \item  ``a DSLR photo of a delicious chocolate brownie dessert with ice cream on the side" 
 \item  ``a DSLR photo of a pair of headphones sitting on a desk" 
 \item  ``a DSLR photo of a quill and ink sitting on a desk" 
 \item  ``a DSLR photo of A very beautiful tiny human heart organic sculpture made of copper wire and threaded pipes, very intricate, curved, Studio lighting, high resolution" 
 \item  ``a DSLR photo of a very cool and trendy pair of sneakers, studio lighting" 
 \item  ``a DSLR photo of a wooden desk and chair from an elementary school" 
 \item  ``a lego tank with a golden gun and a blue flying flag" 
 \item  ``a lego tank with a golden gun and a red flying flag" 
 \item  ``a lego tank with a golden gun and a yellow flying flag" 
 \item  ``a green apple in a hexagonal cup on a round cabinet" 
 \item  ``a green apple on a yellow tray on a red desk" 
 \item  ``a green cactus in a hexagonal cup on a star-shaped tray" 
 \item  ``a green spoon on a red cake in a yellow tray" 
 \item  ``a model of a round house with a spherical roof on a hexagonal park" 
 \item  ``a model of a round house with a spherical roof on a square park" 
 \item  ``a model of a silver house with a golden roof beside a lego man" 
 \item  ``a model of a silver house with a golden roof beside a wooden car" 
 \item  ``a model of a silver house with a golden roof beside an origami coconut tree" 
 \item  ``A candle burns beside an ancient, leather-bound book" 
 \item  ``An apple lays nestled next to a vintage, brass pocket watch" 
 \item  ``A quill pen lies across a stack of unmarked parchment paper" 
 \item  ``A wine bottle and two empty glasses glisten under a chandelier" 
 \item  ``A magnifying glass sits on top of a mysterious, printed map" 
 \item  ``A weathered straw hat hangs beside a freshly picked sunflower" 
 \item  ``An open diary lays flat, a single dried rose on its pages" 
 \item  ``A twinkling star ornament hangs closely with a snow globe" 
 \item  ``A sandy hourglass and a rugged compass lay side by side" 
 \item  ``A pair of spectacles lies open on a dog-eared paperback" 
 \item  ``Ripe apples cluster next to a gleaming knife" 
 \item  ``An abandoned teddy bear leans against a discarded toy car" 
 \item  ``A bottle of red wine stands alongside an empty wine glass" 
 \item  ``A colorful array of spices in tiny jars sits next to an unused cooking book" 
 \item  ``fries and a hamburger

\end{enumerate}

\section{More Qualitative Results}
\label{sec:additional_viz} 

\end{document}